\documentclass[letterpaper,journal]{IEEEtran}

\usepackage{amsmath,amssymb}
\usepackage{booktabs,multirow,array,tabularx}
\usepackage{graphicx}
\usepackage{xcolor}
\usepackage{xspace}
\usepackage{cite}
\usepackage{hyperref}
\usepackage{url}
\usepackage{algorithm}
\usepackage{algorithmic}
\usepackage{tikz}
\usetikzlibrary{arrows.meta,positioning,fit,calc,backgrounds}

\definecolor{stageone}{HTML}{DCEAF7}
\definecolor{stagetwo}{HTML}{E9DDF6}
\definecolor{gamegreen}{HTML}{DDF1E5}
\definecolor{evidenceblue}{HTML}{1F5A94}
\definecolor{strategypurple}{HTML}{7A3E9D}
\definecolor{softgray}{HTML}{F3F4F5}
\definecolor{todored}{HTML}{C62828}

\newcommand{\method}{X-Box\xspace}

\newcommand{\R}{\mathbb{R}}

\newcommand{\Obs}{\mathcal{O}}

\title{X-Box: Hierarchical Cross-Embodiment Boxing for Humanoid Robots}

\author{Kaige Liu$^{1}$, Wentao Dong$^{2}$, Kangning Yin$^{3}$, Yiming Sun$^{4}$, Yang Tang$^{5}$, Yang Li$^{5,*}$, Ming Zhou$^{6,*}$, Weinan Zhang$^{5,*}$
\thanks{$^{1}$Kaige Liu is with the School of Artificial Intelligence and Automation, Huazhong University of Science and Technology, Wuhan 430074, China, and also with Shanghai Innovation Institute, Shanghai 200231, China, and also with Shanghai Artificial Intelligence Laboratory, Shanghai 200232, China.
}%
\thanks{$^{2}$Wentao Dong is with the School of Computer Science,
Shanghai Jiao Tong University, Shanghai 200240, China, and also with Shanghai Artificial Intelligence Laboratory, Shanghai 200232, China. 
}
\thanks{$^{3}$Kangning Yin is with the School of Computer Science,
Shanghai Jiao Tong University, Shanghai 200240, China, and also with Shanghai Innovation Institute, Shanghai 200231, China, and also with Shanghai Artificial Intelligence Laboratory, Shanghai 200232, China.    
}
\thanks{$^{4}$Yiming Sun is with the Guohao School of Tongji University, Shanghai 200092, China, and also with Shanghai Artificial Intelligence Laboratory, Shanghai 200232, China.    
}
\thanks{$^{5}$Yang Tang, Yang Li and Weinan Zhang are with the School of Computer Science,
Shanghai Jiao Tong University, Shanghai 200240, China, and also with Shanghai Artificial Intelligence Laboratory, Shanghai 200232, China.    
}
\thanks{$^{6}$Ming Zhou is with Shanghai Artificial Intelligence Laboratory, Shanghai 200232, China.
}%
\thanks{Yang Li, Ming Zhou and Weinan Zhang are the corresponding authors.}  
}

\begin{document}
\raggedbottom 
\maketitle

\begin{abstract}
High-dynamic robot-to-robot interaction, exemplified by boxing, remains a major challenge for fully autonomous humanoid operation. 
Existing adversarial policies are typically learned via self-play on a single humanoid embodiment.
Although this paradigm has made significant progress, extending such capabilities to cross-embodiment scenarios remains an open challenge due to the diverse morphologies and dynamic constraints of different humanoid embodiments.
To bridge this gap, we present \method, a hierarchical framework that decouples cross-embodiment motion tracking from adversarial strategy learning while sharing a single parameter set across all participating embodiments. 
The framework features a low-level morphology-aware whole-body tracker that injects skeletal topology via a graph neural network to generalize dynamic skills across diverse humanoid embodiments, and leverages a high-level policy-space response oracle (PSRO) module to learn distinct, transferable counter-strategies.
Extensive experiments across four humanoid embodiments show that our method substantially improves cross-embodiment skill transfer and achieves a 97.6\% win rate against the state-of-the-art method. 
Furthermore, real-world deployments on physical A3 and G1 robots confirm the framework's robustness, achieving stable, zero-fall autonomous combat despite real-world sensing occlusions.
Project page: \url{https://sites.google.com/view/x-box-humanoid}.
\end{abstract}



\section{Introduction}
\label{sec:introduction}
Humanoids have recently acquired agile interactive skills such as soccer~\cite{haarnoja2024soccer,wang2026soccer} 
and ping-pong~\cite{ren2026smash}, where the robot reacts to a fast-moving ball or a human partner.
From the perspective of interaction targets, 
these belong to humanoid-object and human-humanoid interaction, 
whose difficulty lies mainly in estimating and controlling object 
states and contact dynamics~\cite{chen2025rhino,su2026hitter,xu2025intermimic}.
Humanoid-humanoid interaction is fundamentally different.
High-dynamic physical 
interaction between humanoids requires continuously perceiving, predicting, 
and responding to the other under coupled actions and 
decision-making relationships that evolve over time~\cite{yin2026robostriker}.

Humanoid boxing is a representative high-dynamic adversarial task. 
Humanoids must execute attacks, defenses, and positional adjustments 
while maintaining whole-body balance and motion stability, 
and respond in real time to the opponent's continuously changing actions. 
This task thus unifies whole-body motion control, physical interaction, 
and adversarial decision-making within a single setting.

\begin{figure}
    \centering
    \includegraphics[width=1.0\linewidth]{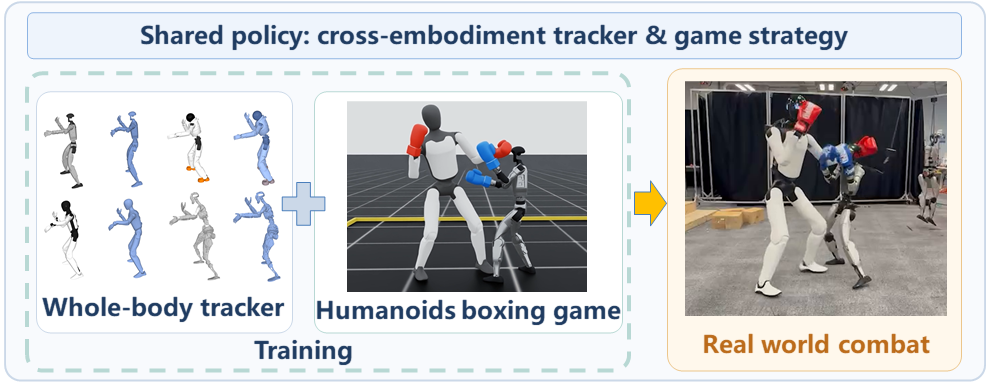}
    \caption{Overview of the proposed framework, which integrates a cross-embodiment whole-body tracker and humanoid boxing game strategies. We train the policy across four humanoid embodiments and deploy it on the Agibot A3 and Unitree G1, with all robots sharing the same model parameters.}
    \label{fig:cross-embodiment-boxing}
    \vspace{-15pt}
\end{figure}

For one single humanoid, whole-body motion control has been
studied in recent years~\cite{liao2025beyondmimic,luo2026sonic,liao2026supermimic},
and has been further extended by diffusion and latent-variable
policies~\cite{yin2025unitracker,zeng2025bfm} and by body-aware
transformer architectures~\cite{sferrazza2025body,zeng2026scalebfm}.
However, most of these methods remain tied to the specific embodiment they are trained on, 
whereas real-world humanoids differ widely in their joint configurations, kinematic properties, and dynamic constraints.
This has motivated cross-embodiment locomotion, including graph-based modular policies~\cite{wang2018nervenet,huang2020smp,kurin2021amorpheus,gupta2022metamorph,hong2022swat,luo2025gcnt},
temporal transformers~\cite{yu2023eat,radosavovic2023humanoidtransformer,liu2025locoformer},
and command-conditioned policies~\cite{ai2025embodimentscaling,lin2025hzero,peng2026eagle,xue2026xhugwbc,doshi2025crossformer}.
Yet their action repertoire is confined to locomotion and producing boxing 
skills such as punches, blocks, and evasive footwork across different morphologies 
requires a cross-embodiment whole-body controller.

However, previous cross-embodiment trackers~\cite{padmanabhan2026venom,zhang2026xwbc,yang2026any2any} condition on morphology externally, 
preventing the temporal network from leveraging skeletal structure. 
We address this by feeding the kinematic graph directly into a GNN, whose output token enables a temporal transformer to produce morphology-invariant whole-body actions.

Even with a cross-embodiment tracker for motion execution, humanoid boxing remains fundamentally challenging due to its adversarial nature: 
the opponent continuously adapts, rendering the environment non-stationary and the continuous strategy space intractable for direct optimization. 
While self-play can foster emergent behaviors~\cite{bansal2018emergent}, it suffers from cyclic dynamics in non-transitive games~\cite{li2023jiangjun}. 
Neural Fictitious Self-Play (NFSP)~\cite{heinrich2016nfsp}, as adopted by RoboStriker~\cite{yin2026robostriker}, mitigates this by averaging past best responses, but inevitably blends mutually countering styles into a single monolithic policy, losing the distinct pure strategies essential for precise counter-attacks. 
Policy-Space Response Oracles (PSRO)~\cite{lanctot2017psro} theoretically resolves this by maintaining a population of separate pure strategies mixed at the meta-level; however, all existing applications of these algorithms are confined to single-morphology agents—either simulated characters~\cite{won2021twoplayer,li2024fightladder} or a single robot platform~\cite{yin2026robostriker}. 
To date, no algorithm supports adversarial strategy learning across heterogeneous humanoid embodiments.
To address this challenge,
we present 
\method,
a hierarchical framework comprising two functionally independent stages
(Fig.~\ref{fig:cross-embodiment-boxing}):
a low-level whole-body controller generalizes dynamic skills across morphologies, 
and a high-level PSRO-based policy population that learns distinct counters transferable among different humanoids.
Both stages share a single set of parameters across all
participating embodiments, enabling truly cross-embodiment
autonomous competition.
This unified architecture allows us to investigate how low-level motion generalization 
and high-level strategic adaptation jointly determine combat performance.

The contributions of this paper are as follows:
\begin{itemize}
    \item \textbf{The first autonomous cross-embodiment humanoid boxing framework,
    validated across simulation and hardware.} 
    \method enables diverse humanoid embodiments to compete
    using a single shared controller and strategy. Simulations
    on four humanoid embodiments demonstrate that \method
    achieves a 97.6\% win rate against the state-of-the-art
    method in over 16,000 simulated bouts and maintains
    zero-fall stable combat in real-world deployment.
    \item \textbf{A morphology-aware whole-body tracker with efficient adaptation.} 
    By encoding skeletal topology via a GNN into a canonical action space, 
    our low-level controller supports simultaneous multi-robot 
    training and rapid post-training transfer to new embodiments, 
    achieving faster convergence and lower tracking error than baselines.
    \item \textbf{A cross-embodiment adversarial strategy learner based on PSRO.} 
    We extend PSRO with relative-size conditioning to decouple strategic 
    reasoning from specific robot dimensions, providing an approximate 
    Nash equilibrium solution in a continuous strategy space that generalizes 
    across heterogeneous platforms.
\end{itemize}

The remainder of this paper is organized as follows.
Section~\ref{sec:problem} formulates the cross-embodiment boxing game and defines the motion-tracking objective.
Section~\ref{sec:method} presents the morphology-injected whole-body tracker and the strategic game strategy.
Section~\ref{sec:experiments} reports tracker transfer,
comparisons with existing boxing methods, component ablations,
and physical deployment.
Section~\ref{sec:conclusion} concludes the paper.

\section{Problem Formulation}
\label{sec:problem}


\subsection{Cross-Embodiment Two-Player Zero-Sum Game}

We study cross-embodiment humanoid boxing as a two-player Markov 
game with 
partial observations~\cite{yin2026robostriker}.
Let \(\mathcal{I}=\{1,2\}\) be the player set and let \(\mathcal{M}\) 
be the set of robot morphologies.
A bout assigns each player a morphology \(m_i\in\mathcal{M}\).
With \(\mathbf{m}=(m_1,m_2)\), the game is
\begin{equation}
M_{\mathbf{m}}
=\bigl\langle
\mathcal{I},\,
\mathcal{S},\,
\Obs,\,
\mathcal{C},\,
P_{\mathbf{m}}
\bigr\rangle.
\label{eq:cross-markov-game}
\end{equation}
Here \(\mathcal{S}\) is the physical state of robots, 
$\mathcal{O}$ the observation space, 
$\mathcal{C}$ the finite set of semantic boxing commands,
and
$P_m:\mathcal{S}\times\mathcal{C}\times\mathcal{C}
\rightarrow \Delta(\mathcal{S})$ the transition dynamics
conditioned on the morphology tuple $m$.
Player \(i\) reads this state as a partial observation \(o_t^{\mathrm{II},i}=h_i(s_t)\in\Obs\), 
which includes the relative positions of the two robots.
\(\mathcal{C}\) is the finite set of semantic boxing commands.
The spaces \(\Obs\) and \(\mathcal{C}\) are shared across morphologies, 
so each player acts by
\begin{equation}
c_t^i\sim\pi_i^{\mathrm{II}}(\cdot\mid o_t^{\mathrm{II},i}),
\label{eq:body-policy}
\end{equation}
where the superscript \(\mathrm{II}\) marks the game policy.
The Stage~I tracker defined below executes these commands
in the transition \(P_{\mathbf{m}}\).
On this interaction, a response to the opponent is trained by 
maximizing the undiscounted shaped return. 
Let $r^{\mathrm{II}}_{i,t}$ denote the shaped reward at timestep $t$ for player $i$. 
The training objective is
\begin{equation}
J_i^{\mathrm{II,sh}}
=
\mathbb{E}
\Biggl[
\sum_{t=0}^{T}
r_{i,t}^{\mathrm{II}}
\Biggr].
\label{eq:player-value}
\end{equation}
The equilibrium uses the zero-sum refereed payoff.
Let
$\tau$ denote a complete bout trajectory. 
The payoff for player $i$ is 
\(J_i^{\mathrm{II}}=\mathbb{E}[v_i(\tau)]\), where \(v_2=-v_1\) and \(v_1(\tau)\in\{+1,0,-1\}\) is the bout outcome.
A Nash equilibrium of this payoff solves
\begin{equation}
\pi_i^{\mathrm{II},\star}
\in
\arg\max_{\pi_i^{\mathrm{II}}}
\min_{\pi_{-i}^{\mathrm{II}}}
J_i^{\mathrm{II}}.
\label{eq:maximin}
\end{equation}

\subsection{Cross-Embodiment Motion Tracking}

Semantic boxing commands are executed by tracking robot-native reference motions.
The active command $c_t \in \mathcal{C}$
selects a reference clip and sets its root-centric goal \(g_t\).
Together with proprioception \(o_t^{\mathrm{prop}}\) and the kinematic graph \(G_m\) of 
the controlled body, the tracker draws a joint residual action
\begin{equation}
a_t^{\mathrm{I}}\sim\pi_\theta^{\mathrm{I}}\bigl(\cdot\mid o_t^{\mathrm{prop}},g_t,G_m\bigr),
\label{eq:track-policy}
\end{equation}
where $a^{\mathrm{I}}_t \in \mathbb{R}^{n_J}$ is the residual
correction on $n_J$ canonical joint positions,
$\pi^{\mathrm{I}}_\theta$ is the tracker parameterized by $\theta$.
Importantly, the semantic command $c_t$ specifies
what skill to execute,
whereas the action $a^\mathrm{I}_t$ specifies how each joint
moves to track the selected reference.
The superscript \(\mathrm{I}\) marks this tracker.
The two players' trackers execute \(c_t^{1}\) and \(c_t^{2}\),
and the physical state moves by
\begin{equation}
s_{t+1}
\sim
P_{\mathbf{m}}\bigl(\cdot\mid s_t,c_t^{1},c_t^{2}\bigr).
\label{eq:stage-interface}
\end{equation}
Let \(r_t^{\mathrm{I}}=r^{\mathrm{I}}(o_t^{\mathrm{prop}},g_t)\) measure the tracking error on that goal.
Discounting this reward over the episode gives the objective of the tracker,
\begin{equation}
J^{\mathrm{I}}(\pi_\theta^{\mathrm{I}})
=
\mathbb{E}
\Biggl[
\sum_{t=0}^{T}
\gamma^{t}
r_t^{\mathrm{I}}
\Biggr],
\label{eq:track-return}
\end{equation}
where $\gamma$ is the discount factor.
The optimal tracker solves
\begin{equation}
\pi_\theta^{\mathrm{I},\star}
\in
\arg\max_{\pi_\theta^{\mathrm{I}}}
J^{\mathrm{I}}(\pi_\theta^{\mathrm{I}}).
\end{equation}


\begin{figure*}[!t]
  \centering
  \includegraphics[width=1.0\textwidth]{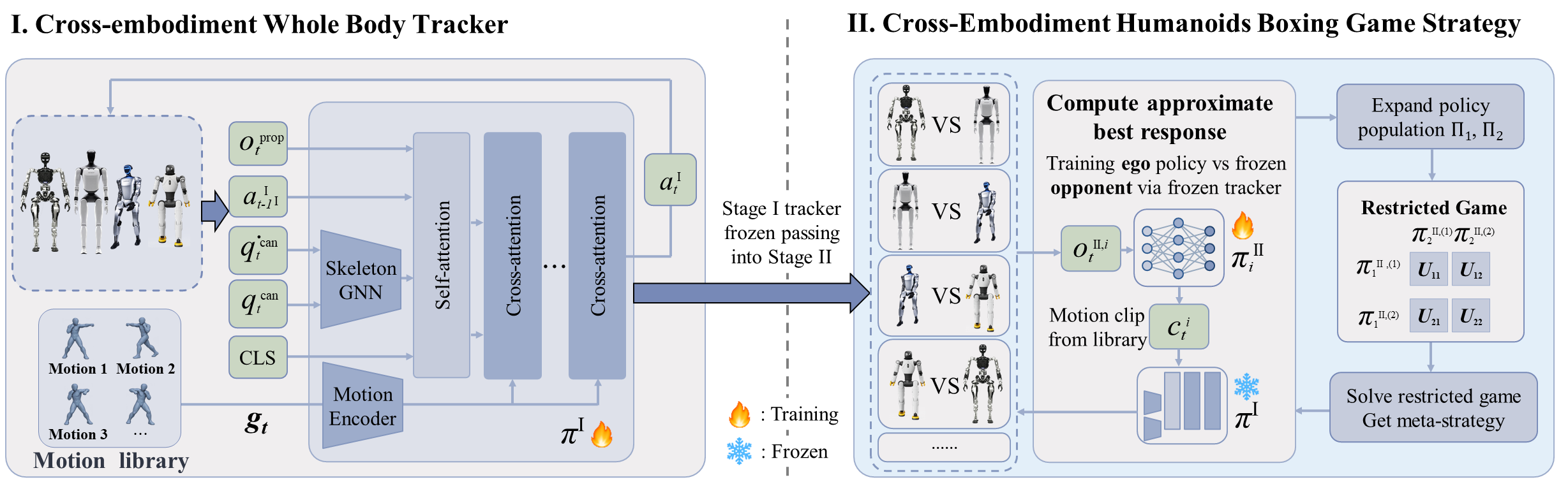} 
  \caption{Overview of the hierarchical framework.   (a): Stage I learns a cross-embodiment whole body tracker. The skeleton GNN encodes canonical joint positions and velocities into a morphology token $\phi_t$. Self attention integrates this token with proprioceptive observations and a learned CLS token, while cross attention incorporates encoded motion references $g_t$ from the reference motion library.   The final CLS representation produces the action $a_t^{\mathrm{I}}$. (b): Stage II learns boxing strategies with the frozen tracker. Policy pairs are evaluated across robot matchups to construct a restricted game with payoff matrix $U$. Solving this game yields meta strategies used to select frozen opponents for training approximate best responses. The resulting policies expand the populations, and the process repeats until the iteration limit is reached or response improvement becomes small.}
  \label{fig:framework}
  \end{figure*}

\section{Methods}
\label{sec:method}
As illustrated in Figure~\ref{fig:framework}, \method decomposes 
cross-embodiment humanoid boxing into two functionally decoupled stages. 
Stage~I trains a single morphology-aware whole-body tracker from human 
motion references, yielding a shared controller applicable to all embodiments. 
Stage~II freezes the trained tracker and employs PSRO to learn a unified 
adversarial policy that generalizes across heterogeneous opponents.

\makeatletter
\let\glove@subsection\subsection
\@secpenalty=10000
\def\subsection{\@startsection{subsection}{2}{\z@}{0.5ex plus 0.2ex minus 0.2ex}%
{0.7ex plus .5ex minus 0ex}{\normalfont\normalsize\itshape}}%
\makeatother
\subsection{Stage I: Cross-Embodiment Whole Body Tracker}
\makeatletter
\let\subsection\glove@subsection
\@secpenalty=-300
\makeatother
\noindent\textbf{Framework Overview and Data Pipeline.} 
Our cross-embodiment tracker employs a Transformer-based policy operating on a unified canonical representation.
At each timestep, the actor takes proprioceptive observations $o_{t-2:t}^{\mathrm{prop}}$, action history $a_{t-2:t-1}^{\mathrm{I}}$, and motion reference goals $g_t$, together with a morphology embedding $\phi_t$ extracted by a skeleton GNN, to predict canonical joint actions $a_t^{\mathrm{I}}$.
Joint states, actions, and link-level motion references follow a shared canonical order across morphologies, while the predicted actions are mapped to morphology-specific joint targets for PD control.

\noindent\textbf{Morphology Encoding via Skeleton GNN.} 
To explicitly capture kinematic structure, we construct a kinematic graph $G_m=(\mathcal{J},E_m)$ for each morphology $m$, where $\mathcal J$ denotes the shared canonical joint set and $E_m$ represents morphology-specific parent-child connections.
Each joint node $u$ is initialized with its canonical position and velocity, $h_u^0=[q^{\mathrm{can}}_{t,u},\dot q^{\mathrm{can}}_{t,u}]$, while link edge features $e^m_{uv}\in\mathbb{R}^3$ encode the zero-configuration relative translation.
We include bidirectional edges and self-loops, with $e^m_{vu}=-e^m_{uv}$ and $e^m_{uu}=\mathbf{0}$.
The GNN employs $L_{\mathrm{g}}=2$ message-passing layers with a hidden dimension of 128 to aggregate joint states and relative link geometry:
\begin{equation}
\begin{aligned}
m_u^\ell &=
\frac{1}{d_u}\sum_{v\in\widetilde{\mathcal N}(u)}
\left(W_h^\ell h_v^\ell+W_e^\ell e^m_{uv}\right),
\end{aligned}
\end{equation}
\begin{equation}
\begin{aligned}
h_u^{\ell+1} &=
\operatorname{ELU}\!\left(
\operatorname{LN}(m_u^\ell+b^\ell)\right),
\end{aligned}
\label{eq:skeleton-message}
\end{equation}
where $\widetilde{\mathcal N}(u)=\mathcal N(u)\cup\{u\}$
and $d_u=|\widetilde{\mathcal N}(u)|$.
The final node features
$H_t=\{h_u^{L_{\mathrm g}}\}_{u\in\mathcal J}$,
are aggregated through mean and max pooling to obtain a compact morphology embedding:
\begin{equation}
\phi_t=\operatorname{Norm}_2\left(
W_p\bigl[
\operatorname{Mean}(H_t)
\Vert\operatorname{Max}(H_t)
\bigr]\right).
\label{eq:morph-token}
\end{equation}
where $W_p\in\mathbb{R}^{64\times256}$ and
$\operatorname{Norm}_2$ denotes $\ell_2$ normalization.
The resulting $\phi_t\in\mathbb{R}^{64}$ encodes both kinematic structure and current joint states.

\noindent\textbf{Motion Reference Encoding.}
To capture short-term tracking targets and longer-term motion intent, we construct $g_t=\{g_{t,k}\}_{k=1}^{K=6}$ from reference frames at offsets $\{0,1,2,3,4\}$ and the last remaining frame.
For each frame $k$, we have
\begin{equation}
g_{t,k}=
\left(
\mathbf{p}_{t,k}^{\mathrm{root}},
\mathbf{p}_{t,k}^{\mathrm{rel}},
R_{t,k}^{\mathrm{root}},
R_{t,k}^{\mathrm{rel}},
\delta_{t,k}
\right),
\label{eq:track-goal-slot}
\end{equation}
where $\mathbf{p}_{t,k}^{\mathrm{root}},\mathbf{p}_{t,k}^{\mathrm{rel}}\in\mathbb{R}^{n_{\mathcal L}\times3}$ denote the reference link positions and their relative displacements from the current tracked links, expressed in the current root frame.
Similarly, $R_{t,k}^{\mathrm{root}},R_{t,k}^{\mathrm{rel}}\in\mathbb{R}^{n_{\mathcal L}\times6}$ represent the corresponding orientations in 6D rotation representation, with $n_{\mathcal L}$ denoting the number of tracked links.
The scalar $\delta_{t,k}$ specifies the temporal offset of each reference frame.
This representation captures the desired motion and its deviation from the current state in a root-relative coordinate system, while the terminal reference frame provides longer-horizon motion guidance.
Each reference frame is encoded into a normalized 32-dimensional motion reference embedding using a shared linear encoder:
\begin{equation}
z_{t,k}=\operatorname{Norm}_2
\left(f_\eta(g_{t,k})\right)
\in\mathbb{R}^{32},
\label{eq:stage1-task-code}
\end{equation} where $f_\eta$ is a shared linear encoder.
The resulting embeddings $z_t=\{z_{t,k}\}_{k=1}^{K}$ provide a compact representation of the reference motion for policy conditioning.

\noindent\textbf{Transformer Context and Policy Output.}
We employ a four-block Transformer with a hidden dimension of 256.
The proprioceptive observation $o_t^{\mathrm{prop}}\in\mathbb{R}^{64}$ comprises projected gravity, base angular velocity, and canonical joint positions and velocities, while $a_t^{\mathrm{I}}\in\mathbb{R}^{29}$ denotes the canonical joint action residual.
Each embodiment is mapped onto this standard skeleton by semantic joint names.
Learned projections $\Phi_o$, $\Phi_a$, and $\Phi_\phi$ map proprioception, previous actions, and morphology embeddings into context tokens, together with a learned $\mathrm{CLS}$ token:
\begin{equation}
\begin{aligned}
X_t=[&\Phi_\phi(\phi_t),\,
\Phi_o(o_{t-2}^{\mathrm{prop}}),\,
\Phi_a(a_{t-2}^{\mathrm{I}}),\\
&\Phi_o(o_{t-1}^{\mathrm{prop}}),\,
\Phi_a(a_{t-1}^{\mathrm{I}}),\,
\Phi_o(o_t^{\mathrm{prop}}),\,
\mathrm{CLS}].
\end{aligned}
\label{eq:stage1-context}
\end{equation}
The motion reference embeddings $z_{t,k}$ are mapped into the shared feature space via a fixed column-orthonormal projection $W\in\mathbb{R}^{256\times32}$.
Each Transformer block performs self-attention over the context tokens, followed by cross-attention to the reference tokens and a SwiGLU feed-forward network, with pre-normalization and residual connections.
Rotary positional embeddings are applied to self-attention, where the $\mathrm{CLS}$ token attends to the full context while remaining masked from other tokens.
A linear head predicts $a_t^{\mathrm{I}}=\Delta q_t^{\mathrm{can}}$ from the final normalized $\mathrm{CLS}$ representation.

\noindent\textbf{Training Objective and Asymmetric Actor-Critic Architecture.}
We train the tracker using PPO with an asymmetric actor-critic architecture, where the critic accesses privileged full-body kinematics.
To preserve motion information in the reference embeddings, an auxiliary decoder $D_\xi$ reconstructs $g_t$ from $z_t$ and detached morphology features:
\begin{equation}
\begin{aligned}
\hat g_t &= D_\xi\left(
\operatorname{vec}(z_t),\operatorname{sg}[\phi_t]\right),\\
\mathcal L_{\mathrm{rec}} &= \frac{1}{\alpha_t^2}
\operatorname{MSE}\left(
\hat g_t,
\operatorname{sg}[\operatorname{vec}(g_t)]
\right),\\
\alpha_t &= \max\left(
\|\operatorname{vec}(g_t)\|_2,1\right),
\end{aligned}
\label{eq:stage1-reconstruction}
\end{equation}
where $\operatorname{vec}$ denotes vectorization, $\operatorname{sg}$ denotes stop-gradient, and $\alpha_t$ normalizes the reconstruction error.
This auxiliary objective updates the motion encoder and decoder while preventing gradients from propagating to the morphology encoder through the reconstruction branch.
The overall training objective combines the standard PPO loss with the auxiliary reconstruction loss:
\begin{equation}
\mathcal L_{\mathrm{total}}
=
\mathcal L_{\mathrm{PPO}}
+\lambda_{\mathrm{rec}}\mathcal L_{\mathrm{rec}},
\label{eq:stage1-objective}
\end{equation}
where $\lambda_{\mathrm{rec}}$ controls the contribution of the reconstruction objective.
The critic shares the actor's four-block Transformer layout but replaces proprioception with privileged kinematics and uses a separate reference encoder. Its reference spans offsets $\{0,1,2,4,8,16,32\}$ with body velocities, encoded into a single 256-dimensional cross-attention token.
After Stage I training, all tracker components are frozen, and only the actor is retained for deployment.

\subsection{Stage II: Cross-Embodiment Humanoids Boxing Game Strategy}

\noindent\textbf{Discrete Command Interface.}
Although the cross-embodiment tracker can follow diverse boxing clips, searching 
in the continuous latent space $z_t$ into which those skills 
are compressed tends to cause mode collapse, so that the policy 
emits only a single behavior \cite{yin2026robostriker}.
We therefore use a discrete semantic command vocabulary.
The high-level policy outputs only a command index,
encoded as the one-hot $c_t^{i,\mathrm{act}}\in\R^{17}$.
These commands are idle, four steps, six punches
(straight, hook, and uppercut on each side), turns to either side,
and four dodges.
The frozen cross-embodiment tracker executes that clip inside the game policy training
loop. Algorithm~\ref{alg:psro} shows the training loop of cross-embodiment humanoids 
boxing game strategy in details. 

\begin{algorithm}[!t]
\caption{Cross-Embodiment Humanoids Boxing Game Strategy}
\label{alg:psro}
\begin{algorithmic}[1]
\STATE \textbf{Input:} frozen cross-embodiment tracker $\pi_\theta^{\mathrm{I}}$; initial populations $\Pi_1,\Pi_2$
\STATE Evaluate $U$ on $\Pi_1\times\Pi_2$ by~\eqref{eq:psro-payoff}
\STATE $(\sigma_1,\sigma_2)\gets\operatorname{PRD}(U,-U)$ 
\FOR{iteration $e=1,2,\ldots,E$}
  \FOR{player $i\in\{1,2\}$}
    \STATE Sample frozen opponents from $\sigma_{-i}$
    \STATE Train oracle $\hat\pi_i^{\mathrm{II}}$ with PPO on $r_{i,t}^{\mathrm{II}}$ through the frozen tracker
    \STATE $\Pi_i\gets\Pi_i\cup\{\hat\pi_i^{\mathrm{II}}\}$
  \ENDFOR
  \STATE Fill missing entries of $U$ by~\eqref{eq:psro-payoff}
  \STATE $(\sigma_1,\sigma_2)\gets\operatorname{PRD}(U,-U)$
\ENDFOR
\STATE \textbf{Output:} $\Pi_1,\Pi_2$ and meta-strategies $(\sigma_1,\sigma_2)$
\end{algorithmic}
\end{algorithm}
\noindent\textbf{Unified Observation Space.}
Let player $i$ face opponent $-i$.
The actor and critic share one observation $o_t^{\mathrm{II},i}\in\R^{38}$.
The field order is
\begin{equation}
\begin{aligned}
o_t^{\mathrm{II},i}=[o_t^{\mathrm{opp},i},\;
(\cos\psi_i,\sin\psi_i),\;
c_t^{i,\mathrm{act}},
\varphi_t^i,\;\beta_t^i,\;x_i],
\end{aligned}
\label{eq:stage2-observation}
\end{equation}
where $o_t^{\mathrm{opp},i}\in\R^{15}$ is opponent perception.
Opponent perception, in meters in $i$'s root frame, comprises 
$R_i^{\top}(\mathbf{p}_{i,w}-\mathbf{p}_{-i,\mathrm{torso}})\in\R^3$ 
and $R_i^{\top}(\mathbf{p}_{-i,w}-\mathbf{p}_{i,\mathrm{torso}})\in\R^3$ for each 
wrist $w\in\{L,R\}$, concatenated left then right into two blocks in $\R^6$, together with the opponent root 
$R_i^{\top}(\mathbf{p}_{-i,\mathrm{root}}-\mathbf{p}_{i,\mathrm{root}})\in\R^3$.
Own heading $(\cos\psi_i,\sin\psi_i)\in\R^2$ is measured in the fixed arena frame.
The command progress is $\varphi_t^i\in[0,1]$, and $\beta_t^i\in\{0,1\}$ indicates that execution has not finished.
Arena position $x_i\in\R^2$ is $i$'s planar root position relative to the arena center.
Every supported morphology uses this field order, so one command policy is shared across bodies.

\noindent\textbf{State-Dependent Action Masking.}
The actor outputs one logit $\zeta_t^i(c)$ per command $c$.
We find that, if we allow the policy to interrupt the current motion,
the policy tends to repeat actions within a single motion slide,
making training difficult.
Thus a state-dependent mask $\chi_t^i$ forbids commands that would interrupt 
an unfinished clip.
Before categorical sampling, masked logits are set to $-\infty$.
While the low-level executor is still running a clip, $\chi_t^i$ retains 
only the active command, so that clip finishes without interruption.
When the executor is free, the policy samples among the remaining legal commands.

\noindent\textbf{Shaped Reward for Response Learning.}
The response oracle is trained with a shaped 
reward $r_{i,t}^{\mathrm{II}}$ combining engagement geometry, 
striking, contact, stability and terminal terms, 
and command regularization, including the configured 
action-diversity reward.
Table~\ref{tab:stage2-reward} gives these terms.
The realized shaped return
$\sum_t r_{i,t}^{\mathrm{II}}$ is the sample of $J_i^{\mathrm{II,sh}}$ 
in~\eqref{eq:player-value}
and is used for response learning and diagnostics.
\begin{table}[t]
\centering
\caption{Terms of the Stage II shaped reward $r_{i,t}^{\mathrm{II}}$.
$\rho_i$ is the planar angle from $i$'s forward axis to the opponent torso.
$\kappa_i$ is the planar distance between the two roots, $\hat{\kappa}_i$ the unit vector from $i$'s root to the opponent root, and $\dot{\mathbf{p}}_{i,\mathrm{root}}$ the planar velocity of the root position $\mathbf{p}_{i,\mathrm{root}}$.
$\kappa_i^{\mathrm{torso}}$ is the planar distance between the two torsos, and $\nu_{i,c}$ counts how often command $c$ has already scored that term.}
\label{tab:stage2-reward}
\scriptsize
\setlength{\tabcolsep}{2pt}
\renewcommand{\arraystretch}{0.92}
\begin{tabular}{@{}l p{0.62\columnwidth} r@{}}
\toprule
Term & Definition & Wt. \\
\midrule
Facing & $\exp(-(1-\cos\rho_i)/0.5)$ & $0.60$ \\
Approach & $\mathrm{clip}(\hat{\kappa}_i^{\top}\dot{\mathbf{p}}_{i,\mathrm{root}},-1.5,1.5)$ if $\kappa_i>0.90$ & $40$ \\
Range & $\exp(-((\kappa_i-0.90)/0.18)^{2})$ & $5$ \\
Crowding & $\max(0,\,0.75-\kappa_i^{\mathrm{torso}})$ & $-60$ \\
Hit & $\mathbf{1}[\text{valid hit}]/\sqrt{1+\nu_{i,c}}$ & $2500$ \\
Outcome & fall/exit: $\mathbf{1}_{\mathrm{opp}}-\mathbf{1}_{\mathrm{own}}$ & $5000$ \\
Diversity & $1/\sqrt{1+\nu_{i,c}}$ if $c$ starts, minus $2$ if repeated & $100$ \\
Switch & $1$ if an unfinished clip is interrupted & $-0.5$ \\
Dodge & successful dodge, $+75$ & $1$ \\
\bottomrule
\end{tabular}
\end{table}

\noindent\textbf{Refereed Payoff and Meta-Strategy.}
Population selection uses independently refereed complete bouts 
rather than the shaped training return.
A scored hit requires fist--torso contact that satisfies force 
and approach-speed conditions, and persistent contact is counted once.
Falls and arena exits can terminate a bout. Otherwise the scored
result at the regulation horizon is the bounded outcome
$v_1(\tau)\in\{+1,0,-1\}$ for player~1, with $v_2(\tau)=-v_1(\tau)$.
The empirical game is this bounded zero-sum outcome.
Win--draw--loss normalization prevents an early lopsided score 
from dominating later population selection.

Let $\mathcal{H}_{\mathrm{train}}$ contain the training matchups.
For policy pair $(p,q)$, evaluate complete bouts under each matchup $h$, with paired resets and swapped arena sides.
First average within each matchup, then weight matchups equally:
\begin{equation}
U_{pq}
=
\frac{1}{|\mathcal{H}_{\mathrm{train}}|}
\sum_{h\in\mathcal{H}_{\mathrm{train}}}
\mathbb{E}_{\tau\sim(\pi_1^{\mathrm{II},(p)},\pi_2^{\mathrm{II},(q)},h)}[v_1(\tau)].
\label{eq:psro-payoff}
\end{equation}
The payoff of player 2 is $-U_{pq}$.
This bounded payoff lies in $[-1,1]$ and equals the matchup-averaged win probability minus loss probability, with draws contributing zero.
Equal matchup weights prevent episode duration or worker throughput from reweighting the game.
Only current population policies and training matchups enter this internal matrix. Held-out pairings never enter population selection.

Decoupled projected replicator dynamics (PRD) computes meta-strategies 
$(\sigma_1,\sigma_2)\in\Delta(\Pi_1)\times\Delta(\Pi_2)$ from the completed 
payoff $U$, where $\Delta(\Pi_1)$ 
and $\Delta(\Pi_2)$ are the probability simplices over the two populations.
At deployment, sample a member from the controlled player's meta-strategy 
at bout start and hold it fixed for that bout.

\section{Experiments}
\label{sec:experiments}
In this section, we evaluate \method through following questions:

\noindent\textbf{RQ1.} Does the cross-embodiment tracker in \method improve transfer efficiency to a new robot?

\noindent\textbf{RQ2.} How does \method compare with existing humanoid boxing methods?

\noindent\textbf{RQ3.} Which components of \method account for its tracking and strategic performance?

\noindent\textbf{RQ4.} How well does \method perform in real-world?



\subsection{Cross-Embodiment Tracker Transfer Efficiency}
\label{sec:tracker-transfer}
\begin{figure}[!t]
\centering
\includegraphics[width=\columnwidth]{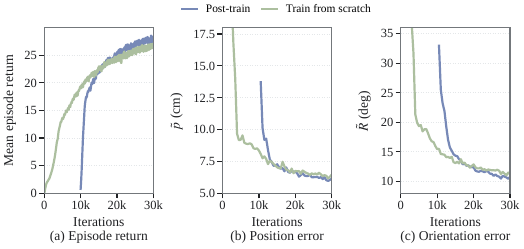}
\caption{Adam Pro training metrics.
(a) Episode return.
(b) Position error.
(c) Orientation error.
Post-training starts from G1, X2, and A3 weights.
The x-axis starts at 10k to include the pretraining updates.
Both curves are trailing-window means through 30k.
All three post-training metrics catch up with and eventually surpass 
those of training from scratch by the end,
meaning cross-embodiment weights speed adaptation to the new robot.}
\label{fig:tracker-transfer-plan}
\end{figure}

To test whether post-training a whole-body tracker for a new robot 
from existing cross-embodiment weights improves training performance on the new robot, 
we run the following experiment.
In this paper, the cross-embodiment tracker contains a 4-layer transformer of width 256.
Motion execution is evaluated by full-clip position error \(\tilde{p}\)
and orientation error \(\tilde{R}\).
We initialize from weights pretrained for 10k iterations on the training sets of 
Unitree G1, Agibot X2, and Agibot A3,
then post-train on PNDbotics Adam Pro using its training set of 38 clips, around 20\,minutes at 50\,Hz.
Figure~\ref{fig:tracker-transfer-plan} shows the mean episode return, position and
orientation errors during training. 
Table~\ref{tab:tracker-transfer-error} reports the error on 8 validation clips, around 6.1\,minutes.
From 20k onward, the post-trained model exhibits lower position and orientation errors 
than the model trained from scratch, demonstrating that cross-embodiment 
pretraining in the proposed tracker accelerates adaptation to the new robot.

\begin{table}[!t]
\centering
\caption{Validation tracking error on 8 clips duration 
training whole-body tracker on Adam Pro.}
\label{tab:tracker-transfer-error}
\footnotesize
\setlength{\tabcolsep}{3pt}
\begin{tabular}{ccccc}
\toprule
& \multicolumn{2}{c}{\(\tilde{p}\downarrow\) (cm)} & \multicolumn{2}{c}{\(\tilde{R}\downarrow\) (deg)} \\
\cmidrule(lr){2-3}\cmidrule(lr){4-5}
Iterations & Post-train & Train from scratch & Post-train & Train from scratch \\
\midrule
20k & \textbf{6.74} & 6.83 & \textbf{12.16} & 12.94 \\
25k & \textbf{6.26} & 6.42 & \textbf{11.36} & 11.58 \\
30k & \textbf{6.07} & 6.53 & \textbf{10.77} & 11.18 \\
\bottomrule
\end{tabular}
\end{table}

\begin{figure}[!t]
\centering
\includegraphics[width=\columnwidth]{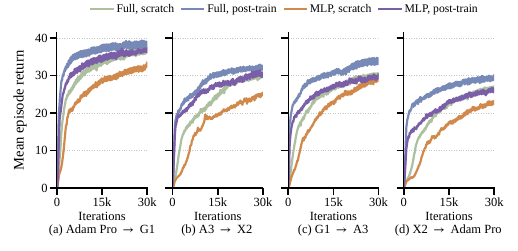}
\caption{Mean episode return for four one-robot transfers:
(a) Adam Pro$\rightarrow$G1, (b) A3$\rightarrow$X2, (c) G1$\rightarrow$A3, and (d) X2$\rightarrow$Adam Pro.
Each panel shows the full tracker and an MLP with no morphology encoder,
both from scratch and post-trained from the source checkpoint 30k iterations.
Post-train stays above scratch for both models,
and the proposed tracker (full) is highest in every panel.}
\label{fig:pair-transfer}
\end{figure}

The post-training gain in Fig.~\ref{fig:tracker-transfer-plan} 
may come from cross-embodiment training.
We therefore pretrain on only one robot and then post-train on a new robot, 
so the model starts without cross-embodiment capability.
If post-training still outperforms training from scratch, 
that capability comes from the network architecture.
Figure~\ref{fig:pair-transfer} compares our architecture with 
a plain MLP on this post-training task.
Both track better after post-training than when trained from scratch, 
and our architecture stays above the MLP in both settings, 
which shows the advantage of our network.

\subsection{Game Strategy with Existing Boxing Methods}

\begin{figure}[!t]
\centering
\includegraphics[width=0.46\textwidth]{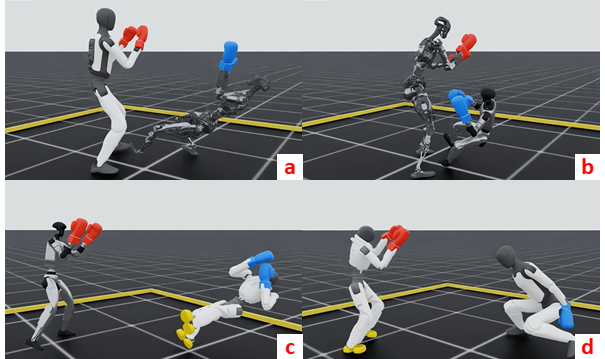}
\caption{Opponent falls in four \method (red glove)--RoboStriker (blue glove) bouts:
(a) A3--Adam Pro, (b) Adam Pro--G1, (c) G1--X2, and (d) X2--A3.
In each frame RoboStriker is down and \method remains standing.}
\label{fig:glove-rs-snapshots}
\end{figure}

To answer RQ2, we compare \method with current humanoid boxing algorithms.
\method trains its PSRO solver for 10 outer iterations.
Each iteration trains one best response per player for 1000 PPO updates and adds it to that player's population.
RoboStriker is the current state of the art in humanoid boxing.
Against the RoboStriker policies, \method wins 97.62\% of 16,384 bouts,
as shown in Table~\ref{tab:rs-matchup-causes}.
Across the 16 ordered matchups, the lowest win rate is 79.2\%, 
and 14 matchups reach at least 97\%.
RoboStriker's average win rate is 2.37\% in all matchups.
Table~\ref{tab:rs-matchup-causes} also shows how each matchup is won 
and lists every cause share and the win rates, with the higher rate in bold.
A draw is left out of both win rates, a mutual fall is counted as a 
self fall and a tied score at the time limit as a lower score, 
so the loss columns equal the opponent win rate plus the draws.
Overall, opponent out-of-bounds and opponent falls account for 
most of \method's wins.
RoboStriker searches in a latent space, so exploration is difficult.
The decoded action must keep balance and play the game at the same time,
and an average policy is fit to the best response, 
which ties the strategy more closely to the opponent it meets.
\method explores in command space.
Its low-level whole-body controller already keeps balance, and the PSRO strategy 
combines efficient exploration with stability and adapts better to a new robot's punching style, 
so its overall win rate is higher.
Figure~\ref{fig:glove-rs-snapshots} shows that in the 
sampled bouts, RoboStriker is the fighter that falls.

\begin{table}[!t]
\centering
\caption{Win rates and cause shares against RoboStriker (\% of 1,024 bouts).
Draws are included in loss columns.
Rounding each entry to 0.01 makes the displayed parts differ from their sum.}
\label{tab:rs-matchup-causes}
\scriptsize
\setlength{\tabcolsep}{1.6pt}
\begin{tabular}{@{}llrr@{\hspace{0.35em}}rrr@{\hspace{0.35em}}rrr@{}}
\toprule
& & \multicolumn{2}{c}{Win rate$\uparrow$} & \multicolumn{3}{c}{Win$\uparrow$} & \multicolumn{3}{c}{Loss$\downarrow$} \\
\cmidrule(lr){3-4}\cmidrule(lr){5-7}\cmidrule(lr){8-10}
\method & Opp. & \method & Opp. & Score & Out & Fall & Score & Out & Fall \\
\midrule
\multirow{4}{*}{G1} & G1 & \textbf{100.00} & 0.00 & 0.00 & 4.10 & 95.90 & 0.00 & 0.00 & 0.00 \\
& X2 & \textbf{100.00} & 0.00 & 0.00 & 3.12 & 96.88 & 0.00 & 0.00 & 0.00 \\
& A3 & \textbf{100.00} & 0.00 & 0.00 & 0.00 & 100.00 & 0.00 & 0.00 & 0.00 \\
& Adam Pro & \textbf{88.18} & 11.72 & 0.00 & 13.09 & 75.10 & 0.00 & 2.25 & 9.57 \\
\midrule
\multirow{4}{*}{X2} & G1 & \textbf{99.61} & 0.39 & 0.00 & 8.30 & 91.31 & 0.00 & 0.00 & 0.39 \\
& X2 & \textbf{100.00} & 0.00 & 0.00 & 2.15 & 97.85 & 0.00 & 0.00 & 0.00 \\
& A3 & \textbf{100.00} & 0.00 & 0.00 & 0.00 & 100.00 & 0.00 & 0.00 & 0.00 \\
& Adam Pro & \textbf{79.20} & 20.80 & 0.00 & 7.32 & 71.88 & 0.00 & 1.46 & 19.34 \\
\midrule
\multirow{4}{*}{A3} & G1 & \textbf{99.80} & 0.20 & 0.00 & 5.57 & 94.24 & 0.00 & 0.00 & 0.20 \\
& X2 & \textbf{100.00} & 0.00 & 0.00 & 1.07 & 98.93 & 0.00 & 0.00 & 0.00 \\
& A3 & \textbf{99.90} & 0.10 & 0.00 & 0.00 & 99.90 & 0.00 & 0.00 & 0.10 \\
& Adam Pro & \textbf{97.95} & 2.05 & 0.10 & 23.34 & 74.51 & 0.00 & 0.49 & 1.56 \\
\midrule
\multirow{4}{*}{Adam Pro} & G1 & \textbf{99.02} & 0.98 & 0.00 & 11.72 & 87.30 & 0.00 & 0.59 & 0.39 \\
& X2 & \textbf{99.71} & 0.29 & 0.00 & 5.96 & 93.75 & 0.00 & 0.20 & 0.10 \\
& A3 & \textbf{100.00} & 0.00 & 0.00 & 0.00 & 100.00 & 0.00 & 0.00 & 0.00 \\
& Adam Pro & \textbf{98.54} & 1.46 & 0.00 & 21.58 & 76.95 & 0.00 & 0.49 & 0.98 \\
\midrule
\multicolumn{2}{@{}l}{All} & \textbf{97.62} & 2.37 & 0.01 & 6.71 & 90.91 & 0.00 & 0.34 & 2.04 \\
\bottomrule
\end{tabular}
\end{table}

\subsection{Component Ablations}
\label{sec:rq5-ablations}
The experiments above treat \method as a complete system. 
This section separates the system into components and tests 
which of them account for the tracking and strategic performance.
We evaluate parameter sharing and architecture at the 
motion-execution layer, then compare \method with a baseline 
that changes only the game solver.
Each comparison holds the remaining interfaces and evaluation 
conditions fixed.

\label{sec:tracker-supporting}

\label{sec:tracker-ablation}

We verify whether the network inside the tracker brings a gain.
To check the skeleton GNN and the transformer backbone in turn, we
replace the transformer with a capacity-matched MLP and
replace the skeleton GNN with a one-hot identity embedding.
Results are shown in Table~\ref{tab:architecture-test-bodies}.
The full transformer with skeleton GNN used by \method has the
lowest position and orientation error on all four robots.

\begin{table}[t]
\centering
\caption{Architecture position and orientation errors on test motions per body.}
\label{tab:architecture-test-bodies}
\footnotesize
\setlength{\tabcolsep}{5pt}
\resizebox{\columnwidth}{!}{%
\begin{tabular}{clrr}
\toprule
Robot & Architecture & \(\tilde{p}\downarrow\) (cm) & \(\tilde{R}\downarrow\) (deg) \\
\midrule
\multirow{3}{*}{G1} & transformer + skeleton GNN (ours) & \textbf{3.24} & \textbf{8.44} \\
 & MLP + skeleton GNN & 4.17 & 10.60 \\
 & transformer + ID embedding & 3.46 & 8.67 \\
\midrule
\multirow{3}{*}{X2} & transformer + skeleton GNN (ours) & \textbf{4.18} & \textbf{9.59} \\
 & MLP + skeleton GNN & 4.73 & 11.72 \\
 & transformer + ID embedding & 4.26 & 10.10 \\
\midrule
\multirow{3}{*}{A3} & transformer + skeleton GNN (ours) & \textbf{4.67} & \textbf{9.39} \\
 & MLP + skeleton GNN & 6.02 & 11.70 \\
 & transformer + ID embedding & 5.13 & 10.08 \\
\midrule
\multirow{3}{*}{Adam Pro} & transformer + skeleton GNN (ours) & \textbf{5.67} & \textbf{10.83} \\
 & MLP + skeleton GNN & 6.27 & 13.28 \\
 & transformer + ID embedding & 6.13 & 10.94 \\
\bottomrule
\end{tabular}
}
\end{table}

\begin{figure}[!t]
\centering
\includegraphics[width=0.46\textwidth]{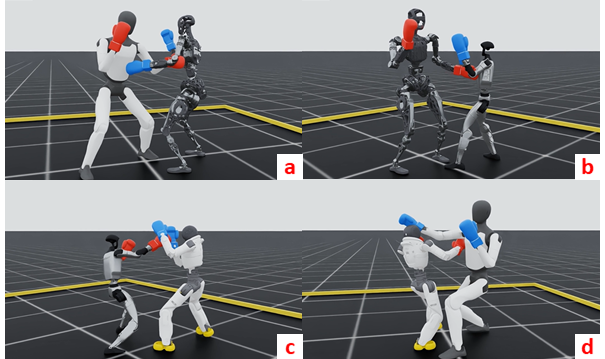}
\caption{Snapshots of \method (red glove) at the moment it throws a punch against the baseline (blue glove):
(a) A3--Adam Pro, (b) Adam Pro--G1, (c) G1--X2, and (d) X2--A3.}
\label{fig:glove-baseline-snapshots}
\end{figure}

RoboStriker does not support cross-embodiment play, 
and it loses nearly every bout against \method.
To test the game solver in \method, we replace that high-level 
solver with NFSP as a baseline, trained at the same
cost as \method.
Table~\ref{tab:baseline-matchup-causes} shows the 16 
ordered matchups between \method and the baseline.
Figure~\ref{fig:glove-baseline-snapshots} shows \method at the moment it throws a punch:
(a) A3--Adam Pro, (b) Adam Pro--G1, (c) G1--X2, and (d) X2--A3.
\method wins more than half the bouts in 13 of the 16 matchups.
On the same-body diagonal, it loses when both sides use G1, 
and wins a majority on X2, A3, and Adam Pro.
\method returns a best response to the population
rather than a best response for each matchup, so the learned
policy can improve the average and still lose some pairings.
Overall, the strategy performs better when it controls A3 or 
Adam Pro than when it controls G1 or X2.
Against RoboStriker, opponent out-of-bounds and opponent 
falls account for most of \method's wins.
In the All row of Table~\ref{tab:baseline-matchup-causes}, 
the Score columns are 40.43\% and 28.77\%, so 69.20\% of the 
16,384 bouts reach the time limit and are decided by score.
The Fall columns in that row are 3.01\% and 3.85\%.

\begin{table}[!t]
\centering
\caption{Win rates and cause shares against the baseline (\% of 1,024 bouts).
Draws are included in loss columns.
Rounding each entry to 0.01 makes the displayed parts differ from their sum.}
\label{tab:baseline-matchup-causes}
\scriptsize
\setlength{\tabcolsep}{1.6pt}
\begin{tabular}{@{}llrr@{\hspace{0.35em}}rrr@{\hspace{0.35em}}rrr@{}}
\toprule
& & \multicolumn{2}{c}{Win rate$\uparrow$} & \multicolumn{3}{c}{Win$\uparrow$} & \multicolumn{3}{c}{Loss$\downarrow$} \\
\cmidrule(lr){3-4}\cmidrule(lr){5-7}\cmidrule(lr){8-10}
\method & Opp. & \method & Opp. & Score & Out & Fall & Score & Out & Fall \\
\midrule
\multirow{4}{*}{G1} & G1 & 46.29 & \textbf{53.61} & 44.14 & 1.86 & 0.29 & 53.12 & 0.20 & 0.39 \\
& X2 & 42.48 & \textbf{57.42} & 22.66 & 5.18 & 14.65 & 56.05 & 0.20 & 1.27 \\
& A3 & \textbf{55.47} & 44.53 & 53.61 & 0.98 & 0.88 & 36.13 & 0.39 & 8.01 \\
& Adam Pro & \textbf{54.30} & 45.61 & 19.34 & 34.38 & 0.59 & 42.19 & 1.76 & 1.76 \\
\midrule
\multirow{4}{*}{X2} & G1 & 33.40 & \textbf{66.60} & 29.20 & 3.71 & 0.49 & 53.32 & 3.61 & 9.67 \\
& X2 & \textbf{53.12} & 46.68 & 40.62 & 9.08 & 3.42 & 41.70 & 2.25 & 2.93 \\
& A3 & \textbf{80.27} & 19.73 & 79.10 & 0.68 & 0.49 & 3.52 & 0.98 & 15.23 \\
& Adam Pro & \textbf{72.66} & 27.25 & 27.64 & 44.24 & 0.78 & 18.07 & 3.42 & 5.86 \\
\midrule
\multirow{4}{*}{A3} & G1 & \textbf{90.43} & 9.57 & 64.36 & 21.68 & 4.39 & 8.59 & 0.98 & 0.00 \\
& X2 & \textbf{68.26} & 31.64 & 46.88 & 18.85 & 2.54 & 29.10 & 2.64 & 0.00 \\
& A3 & \textbf{60.06} & 39.75 & 52.34 & 6.35 & 1.37 & 38.28 & 0.39 & 1.27 \\
& Adam Pro & \textbf{80.18} & 19.82 & 15.62 & 63.38 & 1.17 & 16.11 & 2.05 & 1.66 \\
\midrule
\multirow{4}{*}{Adam Pro} & G1 & \textbf{73.73} & 26.07 & 57.81 & 14.26 & 1.66 & 8.69 & 16.80 & 0.78 \\
& X2 & \textbf{75.88} & 24.12 & 35.94 & 30.27 & 9.67 & 7.62 & 14.75 & 1.76 \\
& A3 & \textbf{61.23} & 38.77 & 42.68 & 14.55 & 4.00 & 31.35 & 2.44 & 4.98 \\
& Adam Pro & \textbf{69.43} & 30.47 & 14.94 & 52.73 & 1.76 & 16.50 & 8.01 & 6.05 \\
\midrule
\multicolumn{2}{@{}l}{All} & \textbf{63.57} & 36.35 & 40.43 & 20.14 & 3.01 & 28.77 & 3.80 & 3.85 \\
\bottomrule
\end{tabular}
\end{table}

\subsection{Real World Deployment}
\label{sec:real}
\begin{figure}[!t]
    \centering
    \includegraphics[width=0.9\columnwidth]{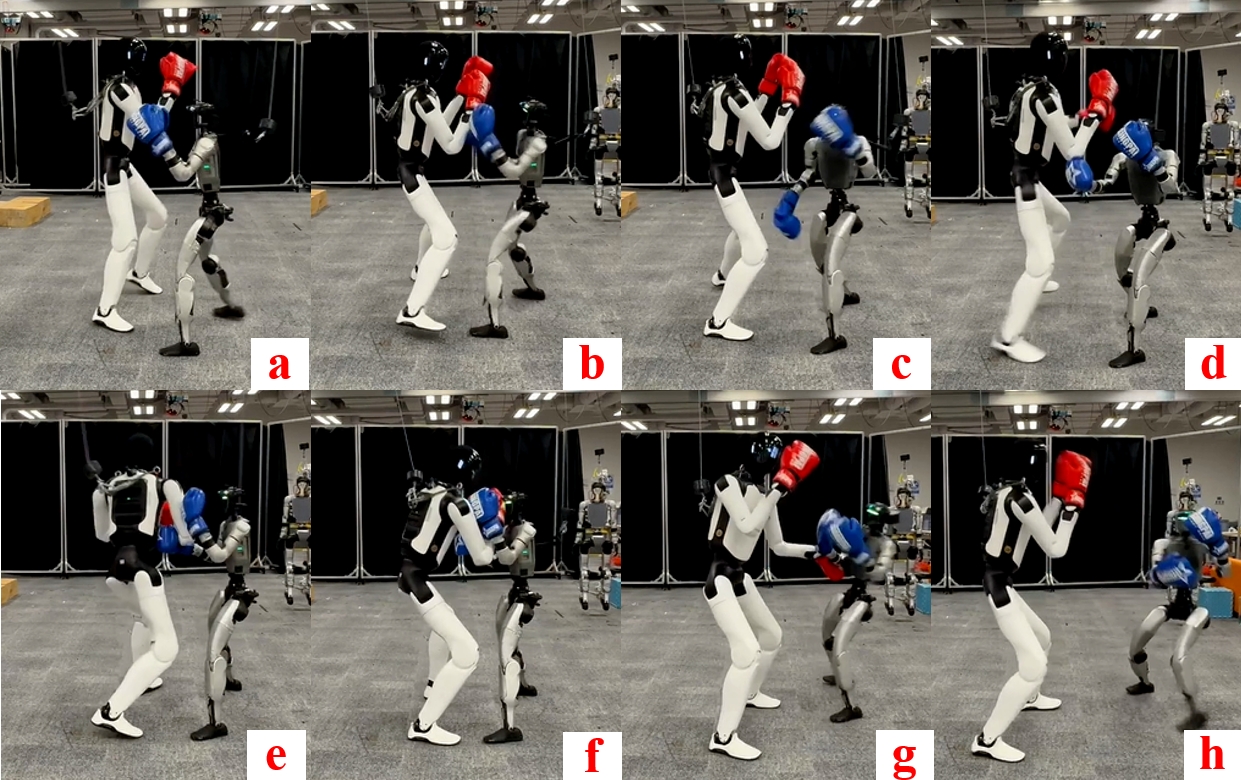}
    \caption{
    Snapshots of an A3 against G1 bout. 
    Panels (a)--(d) show a punch thrown by G1, and (e)--(h) show a punch thrown by A3.
    }
    \label{fig:real}
\end{figure}

We attach an HTC Vive tracker to the back of each robot for 
external global localization.
A kinematic transform maps the tracker position to the robot 
position and yields the robot's global position.
Each robot then combines that position with its own joint 
positions to compute its fist positions and sends them to the opponent.
The receiving robot therefore obtains the opponent information 
in its own ego frame.
We deploy \method and the baseline on the physical A3 and G1 robots,
covering 6 matchups: \method(A3) vs.\ \method(A3), \method(A3) vs.\ \method(G1),
\method(A3) vs.\ baseline(A3), \method(A3) vs.\ baseline(G1), 
baseline(G1) vs.\method(A3), and baseline(G1) vs.\ \method(A3).
Each bout consists of 3 rounds, yielding 18 bouts in total.
Figure~\ref{fig:real} shows consecutive snapshots of 
the A3 and G1 robots punching each other.
Please refer to the supplied multimedia for all the rounds.
Due to the limitations of onboard sensing, 
the robot cannot measure the impact forces it receives. 
Moreover, since none of the robots were knocked down by punches 
in any of the 18 bouts, we instead analyze the behavioral 
discrepancies between simulation and reality.

Table~\ref{tab:real-sim-behavior} compares average time shares in simulation and real-robot bouts.
$|\Delta|$ denotes the mean absolute difference between the four kinds of action shares.
A3 running \method is the closest, with $|\Delta|=7.5$.
On both A3 and G1, the gap comes from punch and idle.
The optical motion-capture system is prone to occlusion while the robots
are moving, causing tracking loss. For safety, our deployed controller
defaults to the idle action whenever the position estimate is lost.
As a result, real-robot behavior cannot fully align with simulation.
The punch rate decreases while the idle share increases.
Nevertheless, the proposed cross-embodiment tracker maintains strong
performance, and no robot fell down in any of the bouts.

\begin{table}[!t]
\centering
\caption{Average time shares of each action within a round, 
in simulation and on the physical robots (\% of live time).
$|\Delta|$ is the mean of $|\mathrm{Sim}-\mathrm{Real}|$ over punch, dodge, move, and idle.}
\label{tab:real-sim-behavior}
\scriptsize
\setlength{\tabcolsep}{2.5pt}
\begin{tabular}{@{}ll rr rr rr rr r@{}}
\toprule
& & \multicolumn{2}{c}{Punch} & \multicolumn{2}{c}{Dodge} & \multicolumn{2}{c}{Move} & \multicolumn{2}{c}{Idle} & \\
\cmidrule(lr){3-4}\cmidrule(lr){5-6}\cmidrule(lr){7-8}\cmidrule(lr){9-10}
Robot & Policy & Sim & Real & Sim & Real & Sim & Real & Sim & Real & $|\Delta|\downarrow$ \\
\midrule
A3 & \method & 43.2 & 29.5 & 12.4 & 14.3 & 7.0 & 6.5 & 35.5 & 49.4 & 7.5 \\
A3 & Baseline & 51.0 & 31.7 & 6.8 & 8.4 & 6.3 & 5.6 & 34.3 & 53.8 & 10.3 \\
G1 & \method & 45.0 & 21.5 & 6.1 & 13.5 & 15.5 & 13.2 & 32.5 & 51.5 & 13.1 \\
G1 & Baseline & 57.0 & 37.3 & 4.5 & 14.1 & 10.6 & 10.8 & 27.4 & 37.8 & 10.0 \\
\bottomrule
\end{tabular}
\end{table}

\section{Conclusion}
\label{sec:conclusion}

This paper presents \method, a hierarchical framework for
cross-embodiment humanoid robot boxing games.
The low level of the framework is a whole-body motion tracker
based on graphs and transformers, and the high level is a
morphology-injected game based on PSRO.
Cross-embodiment pretraining accelerates adaptation to a new
robot and reduces tracking error under the same training budget,
and single-source transfer among the four embodiments yields
positive gains in all four evaluated directions.
Ablation experiments show that both the temporal transformer
and the skeleton GNN contribute to tracking accuracy.
At the game level, \method achieves a higher overall win rate
than the existing boxing system RoboStriker and the NFSP
baseline that also supports cross-embodiment play.

A limitation of this work is that the hierarchically decoupled
control allows the game policy to choose only within a predefined
action space, which restricts further diversification of motions.
In the future, we will extend the framework to more robot
morphologies and a richer command space.


\makeatletter
\global\@colroom=\@colht
\global\vsize=\@colht
\makeatother
\bibliographystyle{IEEEtran}
\bibliography{main}

\end{document}